\documentclass[10pt,twocolumn,letterpaper]{article}
\usepackage{cvpr}

\usepackage{times}
\usepackage{epsfig}
\usepackage{graphicx}
\usepackage{amsmath}
\usepackage{amssymb}
\usepackage{bm}

\usepackage{xcolor,soul}

\usepackage[pagebackref=true,breaklinks=true,letterpaper=true,colorlinks,bookmarks=false]{hyperref}

\cvprfinalcopy 

\def\cvprPaperID{2152} 

\newcommand{\reals}{\mathop \mathbb{R}}

\newcommand{\argmnmz}[1][]{\ensuremath{\arg\;\underset{#1}{\min}\:\:}}
\renewcommand{\vec}{\mathbf}
\newcommand{\Pdv}[2]{\ensuremath{\frac{\boldsymbol{\partial} #1}{\boldsymbol{\partial} #2}}}

\newcommand{\grad}[2]{\nabla_{\!\! #1} #2}
\newcommand{\gradpatch}[3]{\nabla_{\!\! #1}^{#2} #3}

\newcommand{\loss}{\ensuremath{\ell}}
\newcommand{\x}{\vec{x}}

\newcommand{\y}{\vec{y}}

\newcommand{\Z}{\vec{Z}}
\newcommand{\D}{\vec{D}}
\newcommand{\W}{\vec{W}}
\renewcommand{\P}{\vec{P}}

\renewcommand{\a}{\vec{a}}
\renewcommand{\o}{\vec{o}}
\renewcommand{\b}{\vec{b}}
\newcommand{\balpha}{\boldsymbol{\alpha}}

\newcommand{\fL}{^{\text{L}}}

\newcommand{\fSF}{^{\text{SF}}}
\newcommand{\fCSF}{^{\text{CSF}}}

\begin{document}

\title{Sparse Factorization Layers for Neural Networks with Limited Supervision}

\author{Parker Koch and Jason J. Corso\\
Electrical Engineering and Computer Science\\
University of Michigan\\
Ann Arbor, MI 48109\\
{\tt\small \{pakoch|jjcorso\}@umich.edu}
}

\maketitle

\begin{abstract}

Whereas CNNs have demonstrated immense progress in many vision problems, they 
    suffer from a dependence on monumental amounts 
    of labeled training data. On the other hand, dictionary learning does not scale to 
    the size of problems that CNNs can handle, despite being very effective at 
    low-level vision tasks such as denoising and inpainting. Recently, interest 
    has grown in adapting dictionary learning methods for supervised tasks such 
    as classification and inverse problems.  We propose two new network layers 
    that are based on dictionary learning: a sparse factorization layer and a 
    convolutional sparse factorization layer, analogous to fully-connected and 
    convolutional layers, respectively.
    Using our derivations, these layers can be dropped in to existing CNNs, 
    trained together in an end-to-end fashion with back-propagation, and 
    leverage semisupervision in ways classical CNNs cannot.  
    We experimentally compare networks with these two new layers against a baseline CNN.  Our results demonstrate that networks with either of the sparse factorization layers are able to outperform classical CNNs when supervised data are few.  They also show performance improvements in certain tasks when compared to the CNN with no sparse factorization layers with the same exact number of parameters.

\end{abstract}

\section{Introduction}

Artificial neural networks, especially convolutional neural networks (CNN) \cite{LeBoBeIEEE1998} have proven to be powerful, flexible models \cite{HeZhReCVPR2016,SiZiICLR2015,KrSuHiNIPS2012} for a variety of complex tasks, particularly in computer vision \cite{RuDeSuIJCV2015,JoKaFeCVPR2016,OhGuLeNIPS2015}. By composing sequences of simple operations (layers) into complex structures (networks) and then subsequently estimating all network parameters through gradient backpropagation \cite{lecun2012efficient}, they are able to learn representations of complex phenomena, arguably, better than contemporary alternatives.

Despite their widespread adoption and success, these complex networks present 
two core challenges that often make them seem like a black-art to work 
with~\cite{orr2003neural,NgYoClCVPR2015}.  First, they are indeed complex 
structures and hence resist theoretical analysis: the few known results 
\cite{hornik1989multilayer,MaSIAMJC1997} do little to explain what is actually 
being learned or how to improve a certain network.  Second, these complex 
networks have millions of parameters and hence require a large amount of 
supervised data to learn, such as COCO \cite{LiMaBeECCV2014}, ImageNet 
\cite{DeDoSoCVPR2009} or YouTube-8M \cite{youtube8m} which take many 
person-years to build.  This need limits the artificial neural network 
extensibility to situations without large-scale data.  Although some efforts 
toward self-supervision~\cite{Vondrick_2016_CVPR}, which have potential to 
mitigate this issue, have begun, abundant supervision remains a core 
limitation.

On the contrary, sparse representation models~\cite{ElBOOK2010}, like 
dictionary learning through elastic net problem \cite{ZoHaJRSS2005}, afford 
strong theoretical analysis, indeed often rigorous theoretical analysis 
resulting in analytical confidence when used.  They are potentially better 
matched to computer vision, which has many inherently sparse problems.  
Furthermore, due to their structure, they tend to need significantly less training 
data than artificial neural networks.  And, although they have seen success in 
vision, both in generative \cite{YuLiLaCVPR2011} discriminative 
\cite{zhang2010discriminative} forms, at the low- \cite{mairal2008sparse} and 
high-levels \cite{luo2013group}, they have not been able to keep pace with CNNs 
in large-scale or high-dimensional problems \cite{RuDeSuIJCV2015}.


This strong contrast between artificial networks and sparse representations is 
the foundation for our work.  We seek to bridge between these two modeling 
approaches yielding a representation that remains as general as these two are, 
but has better scalability than classical sparse representation learning and, 
at the same time, requires less labeled data than modern artificial neural 
networks.  

Indeed, forging this bridge has been considered by others already.  
For example, greedy deep dictionary learning \cite{TaMaSiarXiv2016} 
sequentially encodes the data through a series of sparse factorizations.  While 
interesting, this approach is neither end-to-end, nor does it afford a 
supervised loss function, limiting its potential in many vision problems.  
Convolutional sparse coding \cite{KaSeBoNIPS2010} derives a convolution form of 
sparse representations, but it is not yet known how to embed these into 
artificial neural networks.  On the other hand, rectified linear units 
\cite{GlBoBeAIStats2011} induce a sparsity in activation, and have improved 
performance.  However, they provide no means for tuning the level of sparsity.   
Summarily, although there has been work in forging this bridge, no approach we 
are aware of successfully unifies sparse representation modeling and artificial 
neural networks with a controllable level of sparsity and a capability to move 
between supervised and unsupervised loss (see Section \ref{sec:related} for more 
discussion).

In this paper, we take a step toward bridging these two disparate fields.
The core novelty in our work is two network layers that do sparse 
representation learning, or dictionary learning, directly within the network.  
We call them sparse factorization layers because they indeed factorize the 
input into its sparse reconstruction coefficients, which are then passed onward 
to later layers in the network---sparse activation.  

We propose both a 
\textit{sparse factorization layer} (SF) that learns the dictionary over all of 
its input terms, analogous to a fully connected layer, and a 
\textit{convolutional sparse factorization layer} (CSF) that slides the 
dictionary over the input analogous (in spirit) to a convolutional layer in a 
CNN.  We show how to compute the back-propagation gradients for the parameters 
(dictionaries) of both layers, leveraging theoretical results on the local 
linearity of dictionaries \cite{MaBaPoTPAMI2012}.  Hence the two layers can be 
incorporated and trained directly within modern networks.  Furthermore, we 
propose a new semisupervised loss that leverages the generative nature of the sparse factorization layers. 

The \textbf{main contribution} of our paper is these two sparse factorization layers that can be plugged into modern convolutional neural networks, trained in an end-to-end manner, and facilitate tunable activation sparsity.  A secondary contribution of the paper is a semisupervised loss function that combines the generative nature of the sparse factorization layers into the discriminative convolutional network.

We demonstrate the empirical utility of the SF and CSF layers as replacements for traditional CNN layers.   
It is beyond the scope of this paper to evaluate all of our stated goals of bridging these two disparate fields.  So, we focus on the use of these two sparse factorization layers with limited data in comparison to a classical convolutional neural network.  On the digit classification problem with MNIST, comparable networks with SF and CSF layers replacing their analogs in the original CNN perform on par with the CNN.  

However, when we reduce the amount of training data to $.167\%$ of the original size (10 samples per class), the performance of our models are $7.39\%$ and $4.23\%$ higher than the CNN, for the sparse factorization and convolutional sparse factorization layers, respectively.  As expected the relative performance improvement of our sparse factorization layers decreases with respect to CNN performance as the amount training data increases.  We also show marked improvement ($1.58\%$ absolute accuracy) in performance on one of the MNIST variations dataset.  Section \ref{sec:experiments} has full results.

\section{Related Work}
\label{sec:related}

We first focus our discussion on methods seeking to bridge between artificial 
neural networks and sparse representation modeling.
Greedy deep dictionary learning \cite{TaMaSiarXiv2016} involves a layered, 
generative architecture that seeks to encode inputs through a series of sparse 
factorizations, which are trained individually to reconstruct the input 
faithfully.  It, however, has no mechanism for end-to-end training and is 
unsupervised, limiting its potential in various tasks like classification, 
given modern performance statistics.  Sparse autoencoders 
\cite{MaFrarXiv2013,HiMomentum2010} do the same using linear operations and 
sparsifying transforms in place of sparse factorization.  Although these models 
can be trained in an end-to-end manner, we are not aware of any work that 
incorporates them into supervised or semisupervised tasks.  In contrast, the SF 
layer that we propose subsumes these capabilities, can be trained end-to-end 
and can be embedded in supervised, semisupervised and unsupervised networks.

Recent work in \textit{truly} convolutional sparse coding 
\cite{KaSeBoNIPS2010,BrErLuCVPR2013} strives to produce a sparse representation 
of an image by encoding the image spatially as a whole, instead of in isolated 
patches, which has been the standard way of incorporating sparse 
representations into image analysis.  This method is similar to our proposed 
convolutional sparse factorization layer: although they embed the sparse coding 
operation in the convolution, we are convolutional only in spirit, choosing to 
map the sparse factorization approach over every patch in the image.  Although 
this work has promise and could potentially even be used to extend the ideas in 
our paper, we are not aware of a result indicating that they can be incorporating 
into artificial 
neural networks, like our methods.

Lastly, hierarchical and multi-layered sparse coding of images \cite{YuLiLaCVPR2011,YaYuHuCVPR2010,BoReFoNIPS2011} has been performed both in a greedy fashion and end-to-end fashion.  However, these typically involve heuristics and lack a principled calculus to embed them within the artificial neural network framework.  They hence lack compatibility with traditional CNN components.  In contrast, both of our new layers can be dropped-in to various artificial (and convolutional) neural networks and trained in the same architecture with back-propagation.

\paragraph{Sparsity in CNNs}
We relate our work to past methods that directly induce sparsity into artificial neural networks.
Various methods \cite{SeNeurComp1997,NgICML2004,BoCuNIPS2008} have indirectly induced sparsity in the parameters or activations by incorporating sparsity-inducing penalties into network loss functions. Techniques like Dropout \cite{SrHiKrJMLR2014} and DropConnect \cite{WaZeZhICML2013} artificially simulate activation sparsity during inference to prevent redundancy and improve performance.
Glorot et al.~\cite{GlBoBeAIStats2011} argue for the use of the rectified linear unit (ReLU), a sparsifying activation function, to improve performance in CNNs. 
This works primarily by limiting the sensitivity of the activations to small changes in the input, thereby forming a more robust and error-tolerant network. However, these methods either lack theoretical basis, offer no means for controlling the activations' sparsity, and/or require that all nonzero activations are positive, unlike our work.

The rest of the paper is organized as follows.  In Section \ref{sec:background} we lay the notational and modeling foundation for the paper.  Section \ref{sec:sparsemain} introduces sparse factorization and the sparse factorization network layers.  Section \ref{sec:experiments} describes our experimental evaluation of our work, and finally, Section \ref{sec:conclusion} summarizes our work and discusses its limitations and future directions.

\section{Background}
\label{sec:background}

\paragraph{Notation}
We have made an effort to maintain consistent notation throughout the paper, sometimes at the expense of deviating from the common notation in existing literature, especially that in dictionary learning.  To that end, we use bold lower-case letters, such as $\x$ and $\a$ to represent tensors that are typically vectors---these are, e.g., inputs and outputs of neural network layers or sparse representation (coding) coefficients.  We use bold upper-case symbols, such as $\P$, to denote tensors that act as operators, such as weight matrices in neural networks and dictionaries in sparse representations (these are generally the parameters we seek to learn).  Greek letters, such as $\lambda$, are used for scalars.

\paragraph{Convolutional Neural Networks}

Artificial neural networks are systems constructed by the composition of potentially many simple, differentiable operations. Let $\a_0 = \x$ denote the input to the network, which is generally a tensor of some order, such as an image.  The first operation, or layer, denoted $f_1$, takes input $\x$ and parameters $\P_1$ to produce an output, or activation, $\a_1$, which serves as input to the second layer.  The process repeats until we reach the activation of the $n$th layer of the network, $\a_n$.  For readability, we use the term activation to refer to the output of each layer in the network, not only those layers commonly called \textit{activation functions} like the sigmoid or tanh layers.

Despite the enormous flexibility of such models, only a small variety of layers are commonly used; most notably, \textit{linear transform layers} ($f\fL(\x;\P)=\P\x$, for vector $\x$ and parameter (weight) matrix $\P$), \textit{convolution layers} ($f(\x;\P,\o)=\x*\P + \o$, for tensor $\x$, kernel bank $\P$, and offset $\o$), and parameter-free transforms like \textit{local pooling}, \textit{rectification}, and \textit{sigmoid} activation.
CNNs are networks with one or more convolution layers.

For any such supervised network, a loss function $\loss_S(\cdot)$ compares the final activation to some supervision information $\y$ about the input $\x$ to measure, for example, classification correctness, or some other measure of output quality. The chain rule is then employed to calculate the gradient of this loss with respect to each intermediate activation: 
\begin{alignat}{3}
\rlap{$\hspace{-0.2cm}\loss_S(\a_n,\y) = L_S$} 
\label{eq:CNN_model}\\
%
\a_n = f_n(\a_{n-1};\P_n) &\quad\nearrow\;\searrow\;&& \grad{\a_{n-1}}{L_S} = \Pdv{f_n}{\a_{n-1}}\grad{\a_n}{L_S}  \nonumber \\
\vdots \quad\quad\quad\quad &\quad\big\uparrow\quad\;\big\downarrow\quad&& \quad\quad\quad\quad \vdots \nonumber \\
\a_2 = f_2(\a_1;\P_2) &\quad\big\uparrow\quad\;\big\downarrow\quad&& \grad{\a_1}{L_S} = \Pdv{f_2}{\a_1}\grad{\a_2}{L_S} \nonumber \\
\a_1 = f_1(\a_0\doteq\x;\P_1) &\quad\big\uparrow\quad\;\big\downarrow\quad&& \grad{\x}{L_S} = \Pdv{f_1}{\x}\grad{\a_1}{L_S}
\enspace.
    \nonumber 
\end{alignat}
\noindent At each step of the backward computation (the right column in (\ref{eq:CNN_model})), the gradient $\grad{\P}{L_S}$ is also computed:
\begin{align}
\grad{\P_i}{L_S} = \Pdv{f_i}{\P_i}\grad{\a_i}{L_S}.
\label{eq:CNN_Pgrad}
\end{align}
With these calculated, gradient descent on the network parameters is used to shrink the overall loss.  
The $L_S$ denotes supervised loss;
we will later incorporate an unsupervised loss into the formulation and denote it by $L_U$.

It should be noted that networks are not limited to a single chain of operations, but we restrict our discussion to such for simplicity and without loss of generality. Additionally, our use of the notation $\boldsymbol{\partial}f/\boldsymbol{\partial}\P$ is purely symbolic, as the parameters $\P$ may be a scalar, vector, matrix, or a collection thereof---or missing altogether.   


\section{Sparse Factorization in Neural Networks}
\label{sec:sparsemain}


Sparse representation models like dictionary learning~\cite{ElBOOK2010} decompose, or 
factorize, a signal into the linear combination of just a few elements from an 
overcomplete basis, commonly called a dictionary.  In contrast, common layers 
in neural networks, such as the linear transform or fully-connected layer, 
compose the output from the parameter matrix and the input.  Concretely, 
consider such a linear transform layer: 
\begin{align}
    \a & = f^{\text{L}}(\x;\P) \nonumber\\
       & = \P\x \enspace,
\end{align}
\noindent for input vector $\x$ and parameters $\P$.  
The (forward) process of sparse factorization is significantly different, involving the solution to an unconstrained optimization problem:
\begin{align}
    \a & = \balpha^*(\x,\P)
\label{eq:sp_coding}
    \\ 
    & = \argmnmz[\hat{\a}] \frac{1}{2}  \lVert \x-\P\hat{\a}\rVert_2^2 +
      \lambda_1\lVert\hat{\a}\rVert_1+\frac{\lambda_2}{2}\lVert\hat{\a}\rVert_2^2
\enspace, \nonumber
\end{align}
\noindent 
where the $\ell_1$-norm is a sparsity-inducing regularizer, and the (typically miniscule) $\ell_2$ penalty on $\a$ is for numerical stability of the optimization algorithm and the $\lambda$ terms are scalar weights on them.    This formulation is referred to as the \textit{elastic-net problem} \cite{ZoHaJRSS2005} and is one variant of many sparse representation formulations; our consideration here is general. 

\begin{figure}[t]
    \centering
    \includegraphics[width=0.8\linewidth]{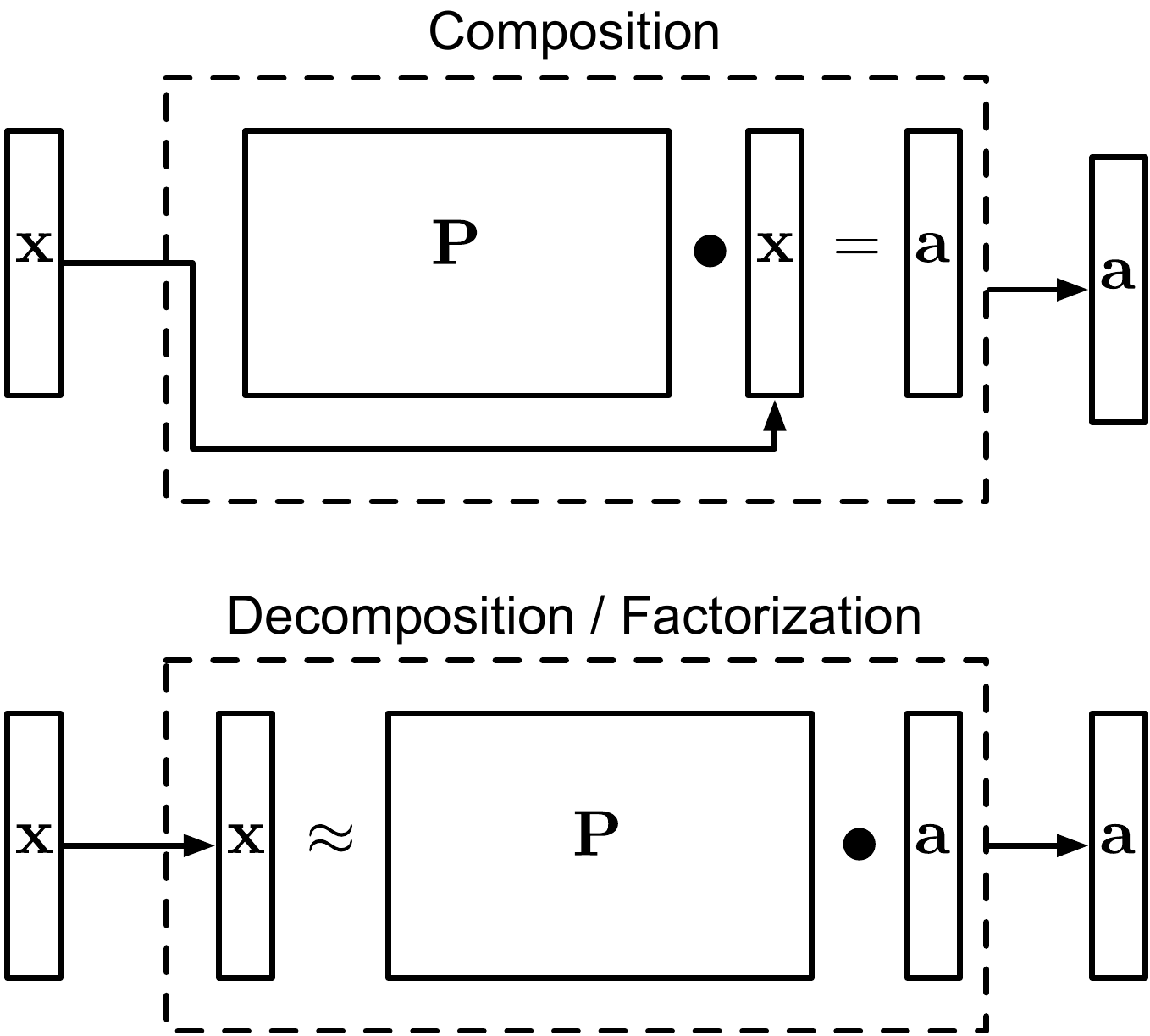}
    \caption{Visual comparison of linear transform operation (composition) and sparse factorization operation (decomposition).  The new layers we propose in this paper are able to induce a sparse activation by leveraging ideas from sparse representation modeling and dictionary learning. $\bullet$ denotes matrix multiplication.}
\label{fig:visual_ltvsf}
\end{figure}

Careful inspection indicates the linear transform takes a dense combination of columns of the parameter matrix $\P$ weighted by the input $\x$ and outputs the results.  However, the sparse representation instead computes the best fit of the input $\x$ with respect to a sparse combination of columns of the parameter matrix $\P$ and then returns the weights of this sparse combination.  In essence, it factorizes $\x$ as a function of $\P$ leading to sparse activations; hence the name of our methods.  
Figure \ref{fig:visual_ltvsf} provides an illustration of this comparison.  
Sparse activations are fundamentally different than the common sparsity inducing regularization used in current neural network training (see Section \ref{sec:related} for more discussion on this relationship).

This difference leads to new neural network layers that are based in sparse representation modeling, yielding sparse activations, strong theoretical guarantees, and the promise of training with less labeled data. 
Indeed, this work builds on the large body of work in sparse representation modeling and dictionary learning that has been prevalent in computer vision 
 \cite{YuLiLaCVPR2011,zhang2010discriminative,mairal2008sparse,luo2013group}.

\begin{figure*}
    \centering
    \includegraphics[width=0.9\linewidth]{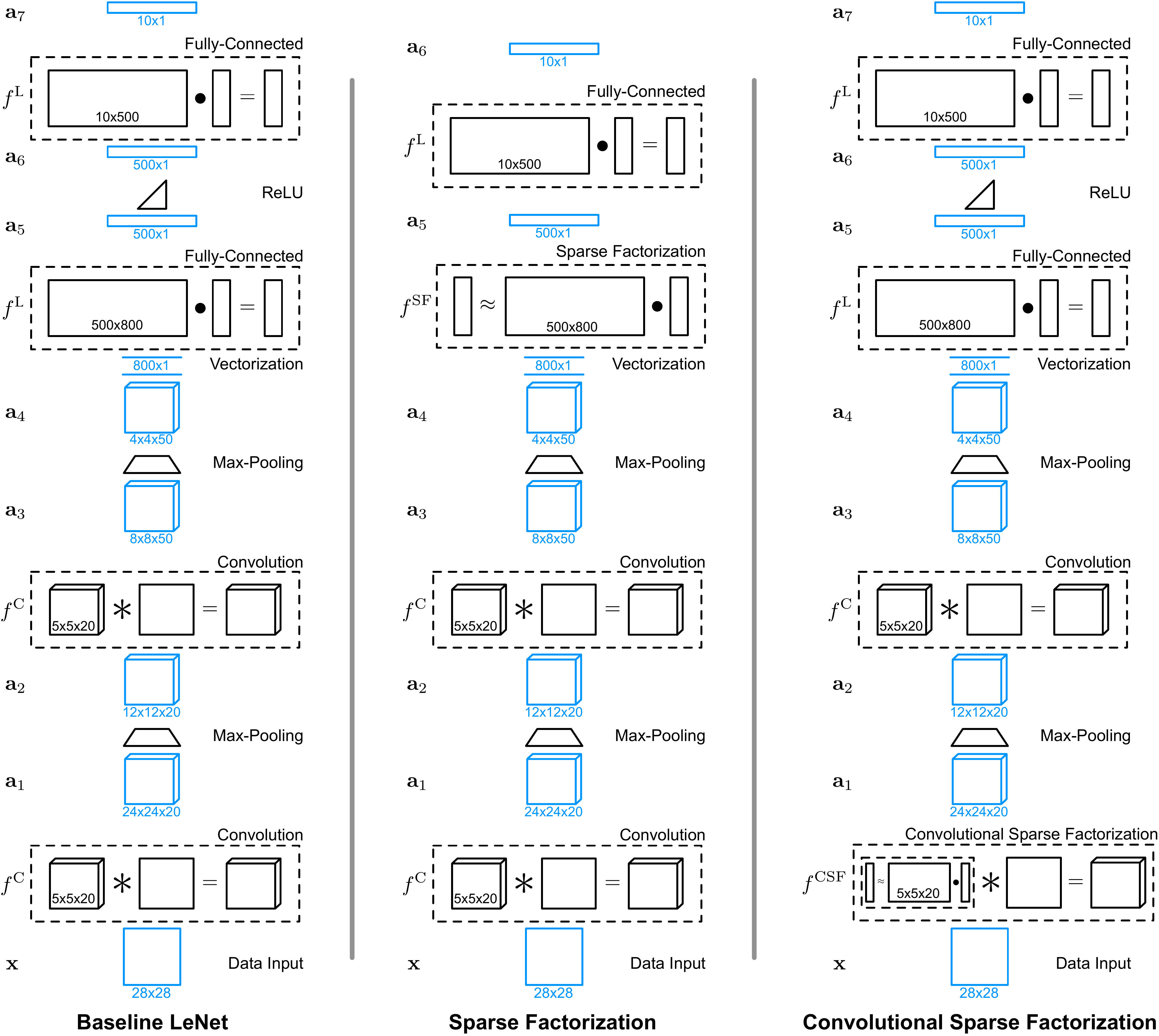}
    \caption{Illustrative comparison of the role our two new sparse factorization layers can provide in a classical network architecture (left) suitable for a digit classification problem like MNIST.  The three networks show the data layer at the bottom and the final activation at the top; computations are drawn in black and activation tensors are drawn in blue. The sparse factorization layer replaces both the fully connected layer and the subsequent rectified linear unit layer (since this would conflict with the possible negative weights of the sparse activation), which is why the SF network has only 6 layers.  The convolutional sparse factorization network replaces the first convolution layer with the CSF layer.  All other things are equal and comparable.   }
    \label{fig:networks}
\end{figure*}

We next show how to create two novel artificial neural network layers based on this core idea of sparse factorization as well as the strong theoretical foundation on which it is based (Section \ref{sec:sparse:theory}).  Examples of the two new layers in comparison to a baseline network based off of LeNet \cite{LeBoBeIEEE1998} is included in Figure \ref{fig:networks}.  We also describe a new semi-supervised loss that leverages the generative nature of sparse representations to regularize the classification loss (Section \ref{sec:semisupervision}). 
%
In Section \ref{sec:experiments}, we describe the results of our experiments using these novel sparse factorization layers, which demonstrate a large gain in performance for limited training data.


\subsection{Sparse Factorization Layer}
\label{sec:sflayer}

The sparse factorization (SF) layer directly implements this sparse representation idea as a new layer that can be plugged into any artificial neural network.  Parameter-for-parameter, the SF layer can replace a linear transform or fully-connected layer in current networks (indeed, we do this in our experiments to make a direct comparison, Section \ref{sec:experiments}).  Forward computation is straightforward, as we will explain, and backward computation leads to the estimation of the parameter matrix (the dictionary) within the gradient descent optimization regime. 

Concretely, a SF layer is parameterized by a dictionary $\P_i\in\reals^{m\times 
K}$, which we assume is known for the forward pass.  The dictionary, or 
parameter matrix, represents the overcomplete basis into which we will 
factorize the input $\a_{i-1}\in\reals^m$ 
\begin{align}
    \a_i & = f\fSF_i(\a_{i-1};\P_i) \label{eq:sflayer}\\
         & = \balpha^*(\a_{i-1},\P_i) \nonumber\\
         & = \argmnmz[\hat{\a}] \frac{1}{2}  \lVert 
         \a_{i-1}-\P\hat{\a}\rVert_2^2 +
      \lambda_1\lVert\hat{\a}\rVert_1+\frac{\lambda_2}{2}\lVert\hat{\a}\rVert_2^2
\enspace,
\nonumber
\end{align}
\noindent where the output activation $\a_i \in \reals^{K}$ is sparse for 
correctly chosen values of $\lambda_1>0$, due to the $\ell_1$-regularizer.  
Again, this is an instance of the elastic-net problem and can be solved with 
Least-Angle Regression (LARS)~\cite{EfHaJoAnnals2004}, completing the forward 
pass of the layer.

During backpropagation, we need to compute the gradient of the loss, $L_S$, with respect to the previous layer's activation $\a_{i-1}$ and the parameter matrix $\P_i$ using the transferred gradient $\grad{\a_i}{L}$ from layer $i+1$. This is possible only if the forward computation procedure is differentiable; Mairal et al.~\cite{MaBaPoTPAMI2012} prove the differentiability of the optimization in $f\fSF(\cdot)$ under mild conditions and compute its gradients, albeit for the problem of task-specific dictionary learning. We present here the results of practical interest to the SF layer computation, with further details in Section \ref{sec:sparse:theory}.

Define $\Lambda$ to be the nonzero support of $\a_i$, i.e., the indices of the coefficients that are nonzero, and define auxiliary variable $\b_i \in \reals^K$ to have $\b_i[\Lambda^C] = 0$ where $\Lambda^C$ is the complement of $\Lambda$ and the $\b_i[\cdot]$ notation means indexing or slicing the vector. Then, let
\begin{align}
    \b_i[\Lambda] = (\P_i^{T}\P_i+\lambda_2\mathbf{I})^{-1} \grad{\a_{i}[\Lambda]}{L_S}
    \label{eq:b}
\end{align}
where $\grad{\a_{i}[\Lambda]}{L_S} = (\grad{\a_{i}}{L_S})[\Lambda]$ is the back-propagated supervised loss from the higher layer indexed at only the nonzero support of $\a_i$ via $\Lambda$.

The gradients of interest are computed directly with this auxiliary vector $\b_i$:
\begin{align}
    \grad{\P_i}{L_S} &= -\P_i\b_i\a_i^{T} + (\a_{i-1}-\P_i\a_i)\b_i^{T}
    \enspace;
    \label{eq:sfgradpi}
    \\
    \grad{\a_{i-1}}{L_S} &= \P_i\b_i
    \enspace.
    \label{eq:sfgradaminus1}
\end{align}
The gradient $\grad{\a_{i-1}}{L_S}$ is backpropagated to lower layers in the network, and the gradient on the parameters, $\grad{\P_i}{L_S}$ is used to update the dictionary $\P_i$ at this sparse factorization layer. Note that after every gradient step, $\P_i$ must be normalized so that each element has $\ell_2$-norm no greater than 1; without this, we could scale $\P_i$ up and $\a_i$ down by a large constant, reducing the regularization penalties arbitrarily without improving the quality of the sparse representation.


%


\subsection{Convolutional Sparse Factorization Layer}
\label{sec:csflayer}

Sparse representations have had success in patch-based models \cite{mairal2008sparse}.  
Likewise, the convolution layer in artificial neural networks has achieved 
great success \cite{HeZhReCVPR2016,SiZiICLR2015,KrSuHiNIPS2012}.  The translation invariance induced by these approaches 
is particularly well-suited to image-based problems.  We hence generalize the 
SF layer to operate convolutionally on image patches.  This new layer, the 
\textit{convolutional sparse factorization} (CSF) layer, performs the 
factorization locally on rectangular input regions much like convolutional 
layers perform localized linear transforms.

Consider such a layer that takes in $h\times w\times c$ images as input and performs the sparse factorization on overlapping patches of size $h_Q\times w_Q$. Let $\a_{i-1}$ be an input image to the CSF layer, and let $\P\in\reals^{h_Qw_Qc\times K}$ be the layer dictionary.  Denote patch $Q$ into image $\a_{i-1}$ as $\a_{i-1}^{(Q)}$ The CSF layer is defined as 
\begin{align}
    \a_i &= f_i\fCSF(\a_{i-1};\P_i) \\
         &= \balpha^*(\a_{i-1}^{(Q)},\P_i) \quad \forall Q \enspace.\nonumber
\end{align}
For each patch $\a_{i-1}^{(Q)}$, a sparse factorization $\a_i^{(Q)}$
is computed in forward inference.  These sparse factorization activation vectors are analogous to the localized patch response to a bank of kernels in traditional CNNs.  These are arranged into a $(h-h_Q+1)\times(w-w_Q+1)\times K$ output tensor--exactly the same size as a traditional convolution layer.  However, note that while we call it a convolutional layer, it is convolutional only in spirit: at each patch location, it is solving a sparse factorization kernel and it uses the same parameter matrix $\P_i$ everywhere.  Defined in this way, the CSF layer is a drop-in replacement for convolutional layers.


During backpropagation, given the output gradients $\grad{\a_{i}}{L_S}$, $\b^{(Q)}$ is computed for each patch as in Eq. \ref{eq:b}.  This allows us to compute patch gradients, $\gradpatch{\P_i}{(Q)}{L_S}$ and $\gradpatch{\a_{i-1}}{(Q)}{L_S}$, for the local dictionary and the current layer activation, respectively adapting Eqs. \ref{eq:sfgradpi} and \ref{eq:sfgradaminus1}.  Due to the linearity of convolution, we simply sum these gradients over all patches to yield the layer gradients that are used to update the parameter (the dictionary) and passed down to the lower layer.

\subsection{Theoretical Basis}
\label{sec:sparse:theory}



To incorporate the sparse factorization and convolutional sparse factorization layers within the end-to-end backpropagation-based training, their respective forward operations need to be differentiable and we need to compute the analytical derivatives.
To that end, we review the pertinent results of Mairal et al. \cite{MaBaPoTPAMI2012}, which serve as the theoretical basis for our two new layers.  They prove
the differentiability of the sparse factorization operation that is used in both $f\fSF(\cdot)$ and $f\fCSF(\cdot)$, and outline an optimization framework that allows sparse coding to be used in a variety of tasks. 

The framework in \cite{MaBaPoTPAMI2012} contains three stages: a vector input 
$\x\in\reals^m$ is first multiplied by the linear transform matrix 
$\Z\in\reals^{p\times m}$; 
given dictionary $\P\in\reals^{p\times K}$,
the sparse code of the resulting vector
is computed through the elastic net, $\balpha^*(\Z\x,\P)$ from Eq. \ref{eq:sp_coding},
%
%
and a 
function $g:\reals^K\to\reals$ is applied to the sparse output, with parameters 
$\W$. Mairal et al.~\cite{MaBaPoTPAMI2012} considered only the cases where $g$ 
is linear or bilinear, though their derivation holds for any differentiable 
$g$, which is critical for our work as will become clear below.

This output, $\W\balpha^*(\Z\x,\P)$, is compared to supervision $\y$ in a twice-differentiable loss function $\loss_S(\y,\W\balpha^*(\Z\x,\P))$, and the task is defined by the minimization thereof. To this end, they prove that $\balpha^*(\Z\x,\P)$ is differentiable for any input $\x$ that has a probability distribution with compact support, which is expected to be satisfied in our operating conditions.  Given this proof, they compute the gradients required to update $\Z$ and $\D$ using a descent algorithm; we have similarly relied on this proof to derive our gradients in Eqs. \ref{eq:sfgradpi} and \ref{eq:sfgradaminus1}.


Although originally developed for compressed sensing, we take a new view onto this framework: it is a shallow artificial neural network.  We detail the forward pass of this shallow network below, using our established convention: 
\begin{align}
\loss_S(\a_3,\y) = L_S & \label{eq:Mairal_model}\\
    \a_3 = g(\a_2;\W) &\quad\big\uparrow \nonumber\\
    \a_2 = f\fSF_2(\a_1;\P) = \balpha^*(\a_1,\P) &\quad\big\uparrow \nonumber\\
\a_1 = f\fL_1(\a_0;\Z) = \Z\a_0 &\quad\big\uparrow \nonumber\\
\a_0 = \x &\quad\big\uparrow
\enspace.
    \nonumber 
\end{align}
Relating this shallow network to the SF and CSF layers in larger networks, we map the last process $g$ as the remainder of the network \textit{above} the inserted SF and CSF layers, noting that their proof holds for any differentiable $g$.  We drop the linear layer and replace it with whatever is \textit{below} the inserted SF and CSF layers without impacting the gradient of the SF or CSF layers in any way.  Also, we have separated the derivation of the backpropagation gradient after splitting the operations into three steps, resulting trivially in Eq \ref{eq:sfgradaminus1}.

\subsection{Semisupervision}
\label{sec:semisupervision}

A fundamental difference between the sparse factorization layers and their traditional neural networks counterparts is the generative nature of sparse factorization.  Consider the extended optimization to include the parameters (dictionary) in addition to the sparse factorization (code), 
\begin{align}
    \P^*,\{\a^*\} &=  \argmnmz[\P,\{\a\}] \sum_j \left[ 
    \frac{1}{2}  \lVert \underset{j}{\x}-\P\underset{j}{\a}\rVert_2^2 +
    \right.
    \label{eq:dictionarylearning}    \\
    &\quad\quad\quad\quad\quad\quad\quad \left.  \lambda_1\lVert\underset{j}{\a}\rVert_1+\frac{\lambda_2}{2}\lVert\underset{j}{\a}\rVert_2^2 \right]
   \nonumber\\
    \text{s.t.} & \quad\lVert \P \rVert_2  \le 1
\enspace, \nonumber
\end{align}
where we use a underset-subscript $\underset{j}{\x}$ to indicate sample index to avoid confusion over layer indices in earlier notation and add the $\ell_2$ regularization on the dictionary.
We immediately notice this goal of a dictionary that facilitates high fidelity reconstruction of the training data.  This goal is unsupervised, in contrast to the supervised goals commonly used to train convolutional neural networks.  Note the relation to auto-encoders \cite{HiSaSCIENCE2006}, which are also generative; however, their layer definitions follow the traditional linear transforms we have already discussed.

Of course, we do not directly solve this non-convex optimization, but solve it instead through backpropagation using the earlier derivations.  However, we do incorporate this goal of seeking a dictionary that enables high fidelity reconstructions of our data, which is similar to multi-task regularization \cite{LaMaSharXiv2016,EvPoICKDDM2004}.  We define the unsupervised loss that can be used to regularize the SF and CSF layer parameter (dictionary) as
\begin{align}
    \loss_U(\a_{i-1},\P_i) = \frac{1}{2}  \lVert \a_{i-1}-\P{\a_i^*}\rVert_2^2 + 
    \lambda_1\lVert{\a_i^*}\rVert_1+\frac{\lambda_2}{2}\lVert{\a_i^*}\rVert_2^2 
\end{align}
\noindent where $\a^*_i$ indicates the activation to the forward pass of SF or CSF layer $i$ after solving the $f\fSF(\cdot)$ or $f\fCSF(\cdot)$, respectively.  This loss can be summed over all samples in a batch.

The gradient associated with this task, $\nabla_\P\loss_U(\a_{i-1},\P)$, is well known in the dictionary learning literature \cite{ElBOOK2010}, though more sophisticated optimization methods are typically preferred for learning purely unsupervised dictionaries. The expression for $\nabla_\P\loss_U$ is given by
\begin{align}
    \grad{\P}{\loss_U} = -(\a_{i-1}-\P\a_i^*){\a_i^*}^T
\enspace.
\label{eq:grad_U}
\end{align}
We can incorporate it into our gradient descent method by defining an effective semisupervised gradient, $\nabla_\P\loss_{SU}$ as a convex combination of the two:
\begin{align}
    \grad{\P}{\loss_{SU}} = (1-\mu)\nabla_\P\loss_S + \mu\nabla_\P\loss_U
\label{eq:loss_SU}
\end{align}
for $0\le\mu\le1$. Note that the sample or activation $\a_{i-1}$ used to calculate the unsupervised loss need not be the same sample or activation used for the supervised loss, nor does it need any corresponding supervision $\y$, allowing for the use of weakly annotated samples.

Finally, because we are layering the sparse factorization layers into larger neural network structures, we need to propagate the gradient from the unsupervised loss down past the dictionary learning for an end-to-end regime.  This gradient,
\begin{align}
    \grad{\a_{i-1}}{\loss_{U}} = \a_{i-1}-\P\a_i^* 
\end{align}
is transferred directly or as part of the $\nabla_{\a_{i-1}}\loss_{SU}$ gradient which is trivially derived.

In practice, we use a step-down schedule for the semisupervision parameter $\mu$, training for several iterations with $\mu=0.8$, then several more with 0.5, 0.3, and finally 0.0. We observe that this improves performance when compared to purely supervised training
This step-down schedule is similar to the one proposed in Mairal et al. \cite{MaBaPoTPAMI2012} except that do not use the piecewise constant-inverse learning rate schedule, favoring instead stochastic gradient descent with a momentum of 0.9.

%
%
%
%
%

\section{Experiments}
\label{sec:experiments}

We evaluate our new layers on a classic vision problem allowing us to thoroughly inspect various aspects of the layers in comparison to a well-studied baseline network.  Although we do expect our new layers to positively impact other vision problems, especially those with limited training data, we restrict the scope of our study to classification at moderate scale.  Recall that Figure \ref{fig:networks} visualizes the three core networks that we will compare.
Our baseline model is a curtailed version of LeNet-5 \cite{LeBoBeIEEE1998}, comprising two convolution layers and two linear layers, and denoted LeNet. Our variants are as follows: CSF replaces the first convolution layer with a CSF layer; SF replaces the first linear layer with a SF layer and removes the rectified linear unit layer; and CSF+SF makes both modifications.  All other things about these networks are equivalent, including the total number of parameters.  

\paragraph{Implementation} All networks are implemented in MatConvNet~ \cite{VeLeACMMulti2015} and trained using stochastic gradient descent with momentum. Forward computation in SF and CSF layers is performed with the SPAMS sparse modeling library~\cite{MaBaPoJMLR2010}.  Please contact the authors for access to the source code.


\subsection{Digit Classification Performance}
We first compare the CSF and SF layers to traditional methods on the task of digit classification. Table \ref{tab:results} shows the networks' classification accuracy on: MNIST \cite{LeBoBeIEEE1998}, a handwritten digit classification dataset; MNIST-rot, a modified MNIST with each digit rotated randomly; MNIST-rand, which contains random noise in the background of every image; and MNIST-img, which superimposes the written digits on randomly selected image patches. The original MNIST has 60000 training images and 10000 for testing and its variants have 12000 for training and 50000 for testing.

We observe improved performance on most tasks through the use of the CSF layer. This is intuitive especially for the MNIST-rand and MNIST-img datasets---on which it outperformed the baseline the most---given the rich history of natural-image denoising and patch reconstruction with sparse representation models~\cite{ElBOOK2010}. The performance of the SF layer is more volatile. This could be due to the unpredictable structure of the network's intermediate activations, which may not readily admit a sparse representation, jeopardizing the stability of the SF layer's output.  Or it could be due to the constraint we imposed to maintain a consistent total number of parameters across the three models being compared; the dictionary underlying the SF layer may not be \textit{wide} enough.

\begin{table}[]
\centering
\caption{Classification accuracy scores on the digit classification datasets, in percent. Boldface indicates the best score for each dataset.}
\label{tab:results}
\begin{tabular}{|r|c|c|c|c|}
	\hline
	           & LeNet      & CSF        & SF    & CSF+SF \\\hline\hline
	MNIST      & 99.14      & \bf{99.20} & 98.31 & 98.40  \\ \hline
	MNIST-rot  & \bf{91.84} & 91.76      & 78.63 & 75.49  \\ \hline
	MNIST-rand & 95.11      & \bf{95.34} & 93.33 & 90.83  \\ \hline
	MNIST-img  & 91.84      & \bf{93.42} & 89.42 & 88.77  \\ \hline
\end{tabular}
\end{table}

\subsection{Effect of Training Data Volume}

\begin{figure}[t]
	\begin{center}
		\includegraphics[width=0.95\linewidth]{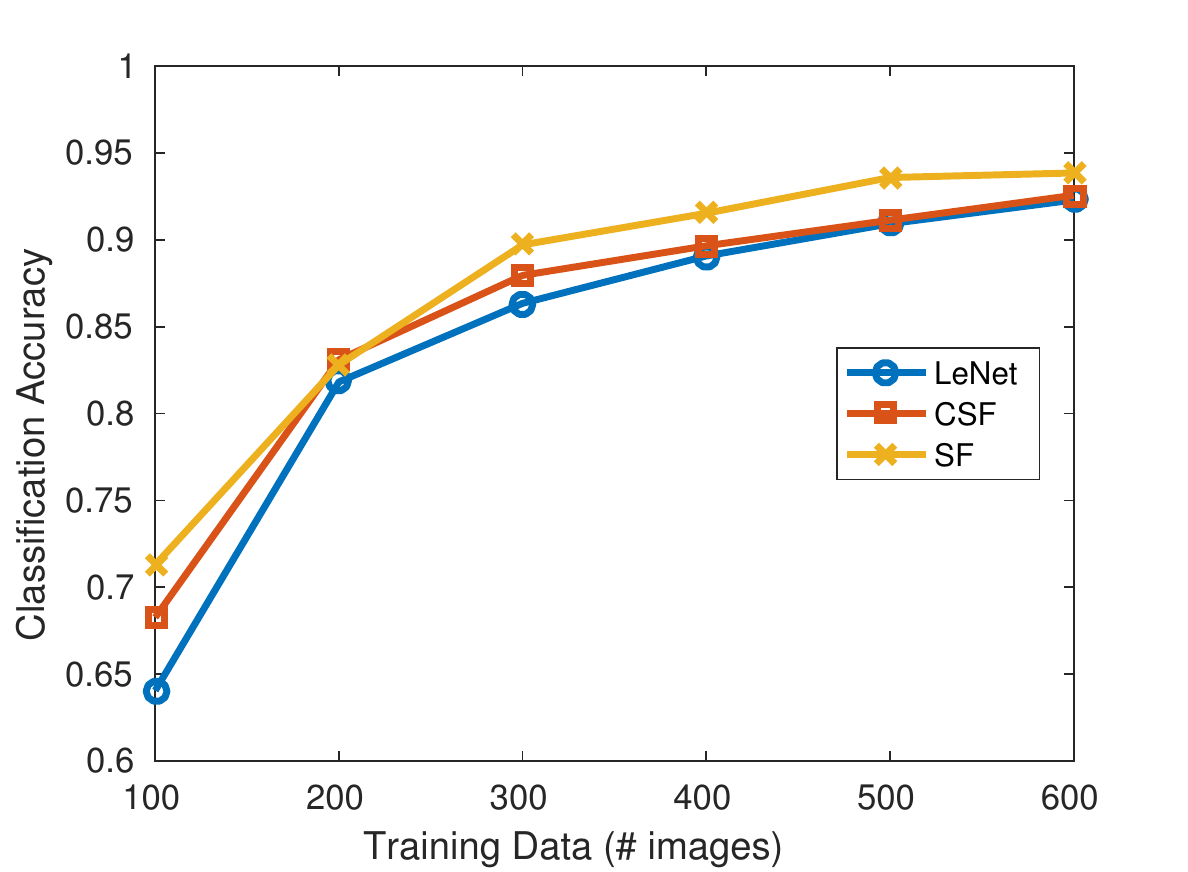}
	\end{center}
	\caption{Classification accuracy of the baseline LeNet, CSF, and SF networks for limited amounts of training images, averaged over several trials. Error bars indicate standard deviation.}
	\label{fig:trainsize}
\end{figure}

To explore our original hypothesis that CNNs with sparse factorization layers will perform comparatively better than traditional CNNs in the face of limited supervised data, we also evaluate our layers' performance on very small training data sets. By varying the number of training samples from 600 (1\% of total training set) to 100 (0.167\%), we observe how each network deals with severely limited available supervision. Each data point in Figure \ref{fig:trainsize} is the mean of several trials of each method, where each trial is trained on a unique subset of the data.

The networks containing sparse factorization layers outperform the baseline on limited training data, with the accuracy gap widening as data becomes scarcer, suggesting they may be more resistant to overfitting.  When supervision is most scarce (100 samples in this experiment), the SF and CSF networks outperform the LeNet network by $7.39\%$ and $4.23\%$, respectively. 
Surprisingly, the SF layer outperforms both the baseline and CSF network despite underperforming them when trained on the full dataset, as shown in Table \ref{tab:results}. This may be an artifact of our na{\"i}ve network design, or an indication that the structure of the sparse factorization problem may be inherently amenable to learning in low-supervision settings.

\section{Conclusions}
\label{sec:conclusion}
In this paper, we have presented two novel CNN layers based on sparse representational modeling in contrast to--and in tandem with--more traditional, compositional layers.  The two layers, sparse factorization and convolutional sparse factorization layers, analogous to fully-connected and convolutional layers in traditional CNNs, are similarly able to be dropped into existing networks.  They are trained with end-to-end back-propagation as well, but they can also be trained with a semisupervised loss, given their origins in unsupervised learning.  Our experiments clearly demonstrate potential in networks with these layers to train strong networks with limited labeled data.

\paragraph{Limitations and Future Work}
Much remains to be explored with the SF and CSF layers. Solving an optimization problem in every forward pass is expensive compared to traditional CNN units, and our implementations typically exhibited training speeds 10 to 40 times slower than with the pure CNN (although they are implemented only on the multicore CPU and not GPU). We postulate that this can be offset by the use of pretraining; since both factorization layers are capable of producing and propagating an unsupervised loss, they (and layers prior to them) can be trained in a wholly unsupervised fashion to initialize the network parameters. The factorization layers also present unique challenges in designing the network that we hope to address in future work, such as choosing the placement, width (dictionary size), and sparsity hyperparameters for CSF and SF layers, all of which we observed have a non-negligible effect.  

\paragraph{Acknowledgements} This work was partially support under ARO W911NF-15-1-0354.


\end{document}